\pdfoutput=1
\documentclass[11pt]{article}
\usepackage[]{acl}
\usepackage{times}
\usepackage{latexsym}
\usepackage[T1]{fontenc}
\usepackage[utf8]{inputenc}

\usepackage{microtype}

\usepackage{graphicx}
\usepackage{tabularx}
\usepackage{multirow}
\usepackage{booktabs}
\usepackage{amsmath}
\usepackage{bbold}
\usepackage{bm}
\usepackage{relsize}
\usepackage{subcaption}

\newcommand{\astfootnote}[1]{
    \let\oldthefootnote=\thefootnote
    \setcounter{footnote}{1}
    \renewcommand{\thefootnote}{\fnsymbol{footnote}}
    \footnotetext{#1}
    \let\thefootnote=\oldthefootnote
}

\title{Probing Out-of-Distribution Robustness of Language Models with Parameter-Efficient Transfer Learning}

\author{Hyunsoo Cho$^\dagger$, Choonghyun Park$^\dagger$, Junyeop Kim$^\dagger$, Hyuhng Joon Kim$^\dagger$, \\ 
        \textbf{ Kang Min Yoo$^\dagger$$^\ddagger$$^*$, Sang-goo Lee$^\dagger$$^\mathsection$$^*$}\\
        $^\dagger$ Seoul National University, $^\ddagger$ NAVER, $^\mathsection$ IntelliSys\\
        \texttt{\{johyunsoo,pch330,juny116,heyjoonkim,sglee\}@europa.snu.ac.kr} \\ 
        \texttt{\{kangmin.yoo\}@navercorp.com}
        }

\begin{document}
\maketitle

    \begin{abstract}
    As the size of the pre-trained language model (PLM) continues to increase, numerous parameter-efficient transfer learning methods have been proposed recently to compensate for the tremendous cost of fine-tuning.
    Despite the impressive results achieved by large pre-trained language models (PLMs) and various parameter-efficient transfer learning (PETL) methods on sundry benchmarks, it remains unclear if they can handle inputs that have been distributionally shifted effectively.
    In this study, we systematically explore how the ability to detect out-of-distribution (OOD) changes as the size of the PLM grows or the transfer methods are altered.
    Specifically, we evaluated various PETL techniques, including fine-tuning, Adapter, LoRA, and prefix-tuning, on three different intention classification tasks, each utilizing various language models with different scales.
    \end{abstract}
    
    \astfootnote{Co-corresponding authors.}
    \section{Introduction}
        Pre-trained language models (PLM), which are pre-trained on large-scale corpora using transformer-based architectures \cite{vaswani2017attention}, have achieved groundbreaking success on sundry benchmarks \cite{wang2018glue,rajpurkar2016squad, wang2019superglue}, establishing themselves as the standard neural model in countless applications.
        Moreover, language models pre-trained with larger parameters on a rich volume of corpora tend to exhibit more intriguing potentials, such as the ability to capture world knowledge \cite{petroni2019language}, generate codes \cite{poesia2022synchromesh}, and even solve mathematical problems \cite{henighan2020scaling}, on top of understanding linguistic knowledge (e.g., semantic or syntactic).
        To explore the apex of pre-trained language models (PLMs), the size of PLMs is growing exponentially and has reached billions to a trillion \cite{brown2020language, chowdhery2022palm, fedus2021switch, hoffmann2022training}.
        
        Under these circumstances, the conventional method for transferring PLMs to a target task (i.e., fine-tuning) is now infeasible as it entails prohibitive costs to train and store the entire parameters of large PLMs for every desired task.
        To mitigate this issue, several recent parameter-efficient transfer learning (PETL) methods have been proposed to improve task scalability. 
        For instance, adapter-based \cite{houlsby2019parameter, hu2021lora} approaches insert small neural modules into each layer of the PLM and update those lightweight modules in the training phase.
        Inspired by the recent success of textual prompts \cite{brown2020language}, prompt-based methods \cite{li2021prefix, lester2021power, shin2020autoprompt} concatenate extra tunable tokens to the front of the input or hidden layers and update prepended soft prompts in the training phase.
        
        Despite these breakthroughs in NLP, even very recent anomaly detection studies \cite{cho2022enhancing, shen2021enhancing} are still limited to relatively small bi-directional PLMs (e.g., BERT, RoBERTa).
        Thus, \textit{how large-scale PLMs or auto-regressive PLMs cope with outliers} is uncharted territory, naturally begging the following questions:
    
        \noindent$\bullet$ \textbf{Q1}: Does increasing model size improve OOD detection performance without model parameters?
    
        \noindent$\bullet$ \textbf{Q2}: If so, does scaling the size of PLM makes the model robust enough to utilize them without any additional process?
    
        \noindent$\bullet$ \textbf{Q3}: Do fine-tuning and various PETL methodologies display differences in OOD detection performance according to the size of PLMs?
    
        \noindent$\bullet$ \textbf{Q4}: Can the OOD detection methods from previous works (usually for the bi-directional PLMs) be transferred to auto-regressive PLMs (GPT)?

        To resolve these questions, this paper investigates the capability of large PLMs as outlier detectors from various perspectives.
        Specifically, we compare the robustness to outliers with various transfer learning techniques on several OOD benchmarks: Full fine-tuning, LoRA \cite{hu2021lora}, Adapter \cite{houlsby2019parameter}, and prefix-tuning \cite{li2021prefix} on various auto-regressive PLMs with different sizes, i.e., GPT2-S, M, L, XL \cite{radford2019language}, GPT-Neo \cite{gpt-neo} and GPT-J \cite{gpt-j}.
        From in-depth investigations, we share several intriguing observations:
        (1) As the size of the PLM increases, the performance improves without any update of model parameters. However, it is still challenging to use it without supervision since their performances still lag far behind compared to the fine-tuned small PLM (i.e., BERT-base).
        (2) PETLs outperform fine-tuning with sufficiently large PLMs in both IND and OOD metrics.
        (3) Lastly, leveraging the information of the last hidden representation, which is the most prevailing method for bi-directional PLM in recent OOD detection, does not transfer well in auto-regressive PLM, requiring a novel representation extracting technique.
        We believe that these findings will help future anomaly detection studies.

    \section{Probing OOD Robustness}
        \subsection{Backbones and Models}
            To investigate the trend of OOD performance under varying scales of PLM, we consider three factors during backbone selection.
            They should be (1) publicly available, (2) reasonably large, and (3) share identical structures to eliminate factors other than size. 
            Since recent large PLMs utilize auto-regressive objectives due to their computational complexity, we adopt six auto-regressive PLMs as the backbone of our experiments accordingly: \textbf{GPT2 (S,M,L,XL)}, \textbf{GPT-Neo}, and \textbf{GPT-J}.
            
            For the parameter-efficient transfer methods, we selected two methods: two adapter-based and one prompt engineering-based.
            Namely, \textbf{Adapter} \cite{houlsby2019parameter}, \textbf{LoRA} \cite{hu2021lora}, and \textbf{Prefix-tuning} \cite{li2021prefix} are selected for the adapter approach, which is compatible with classification tasks, for the prompt approach. 
            We also report the performance of linear evaluation, i.e., single layer perceptron (SLP) on top of PLMs, and fine-tuning, which act like a lower-bound and upper-bound, respectively.
            
        \subsection{Dataset and Metrics}
            \label{sec:dataset}
            \textbf{Dataset.} 
            We evaluate our model on two datasets, CLINC150 and Banking77, widely used in OOD detection.
            CLINC150 dataset \cite{larson2019evaluation} contains 150 class labels (15 intents for 10 domains), while Banking77 dataset \cite{casanueva-etal-2020-efficient} consists of fine-grained 77 bank-related intents.
            Following the experimental settings from previous works \cite{cho2022enhancing, zhang2021pretrained, shu2017doc, fei2016breaking, lin2019deep}, we validate our models in two different scenarios: far-OOD setting and close-OOD setting.
            For CLINC dataset, we train our model with the whole training dataset and test with an independent OOD test split from CLINC dataset, which does not overlap with 150 classes in the training dataset.
            Outliers in CLINC OOD split are distributionally far from the training distribution \cite{zhang2021pretrained}, so it is relatively easy to discern.
            For Banking77, we partition the dataset into 2 disjoint datasets (i.e., IND / OOD dataset) based on the class label.
            Since both IND and OOD datasets originated from the equivalent dataset, they share similar distributions and properties, making the task more demanding.
            Thus, we refer to a CLINC OOD setting as far-OOD and split settings in Banking as close-OOD settings, respectively.
        
            \noindent \textbf{Metrics.} To evaluate IND performance, we measured the classification accuracy.
            And for OOD performance, we adopt two metrics commonly used in recent OOD detection literature:\\
            \noindent$\bullet$ \textbf{FPR@95.} The false-positive rate at the true-positive rate of 95\% (FPR@95) measures the probability of classifying OOD input as IND input when the true-positive rate is 95\%. \\
            \noindent$\bullet$ \textbf{AUROC.} The area under the receiver operating characteristic curve (AUROC) is a threshold-free metric that indicates the ability of the model to discriminate outliers from IND samples.

    %--- figure start---
    \begin{figure*}[t]
    \centering
        \begin{subfigure}{.49\linewidth}
            \centering
            \includegraphics[width=.99\linewidth]{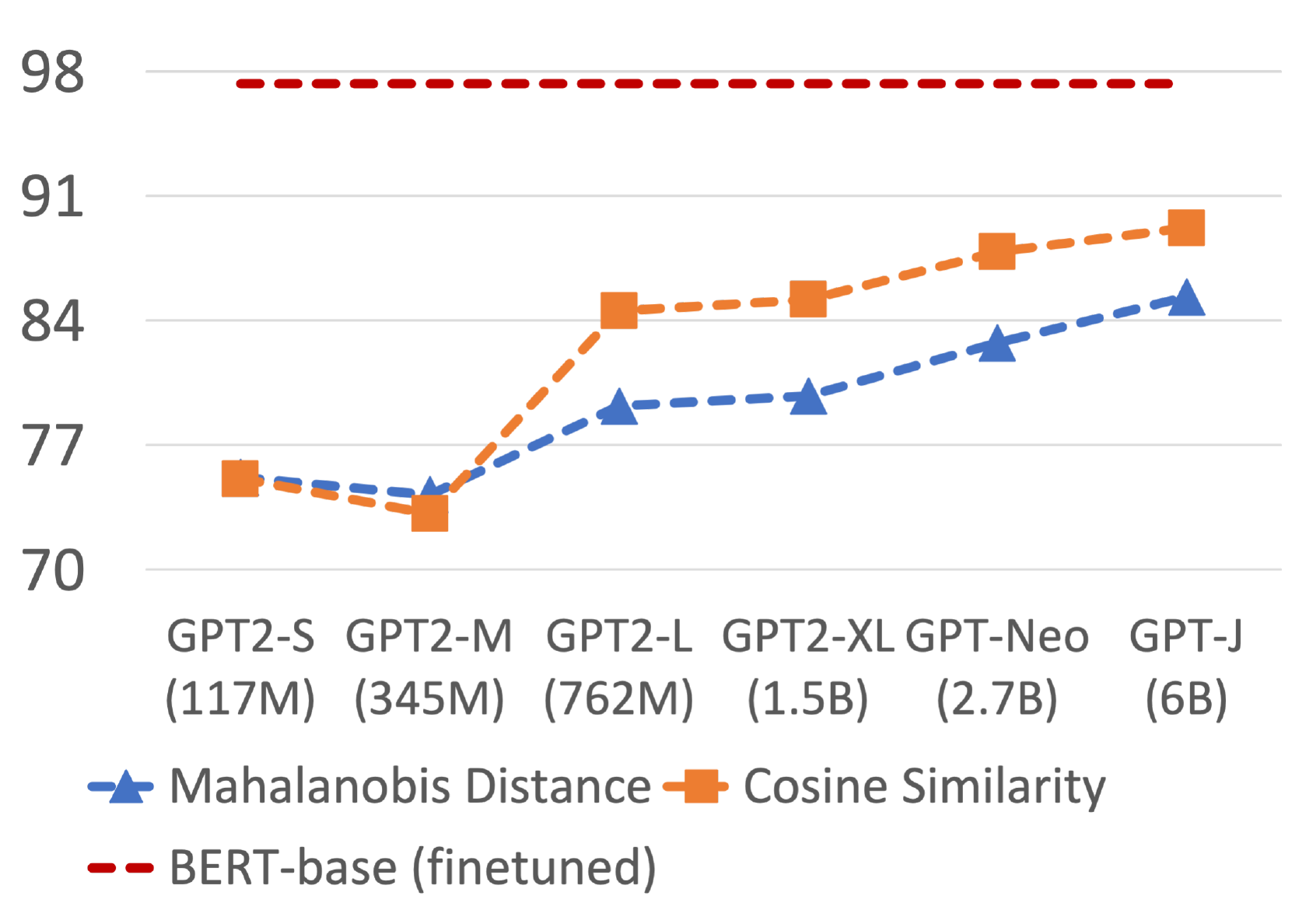}
            \caption{Performance on far-OOD setting.}
            \label{fig:how-zeroshot-far-ood}
        \end{subfigure}
        \begin{subfigure}{.49\linewidth}
            \centering
              \includegraphics[width=.99\linewidth]{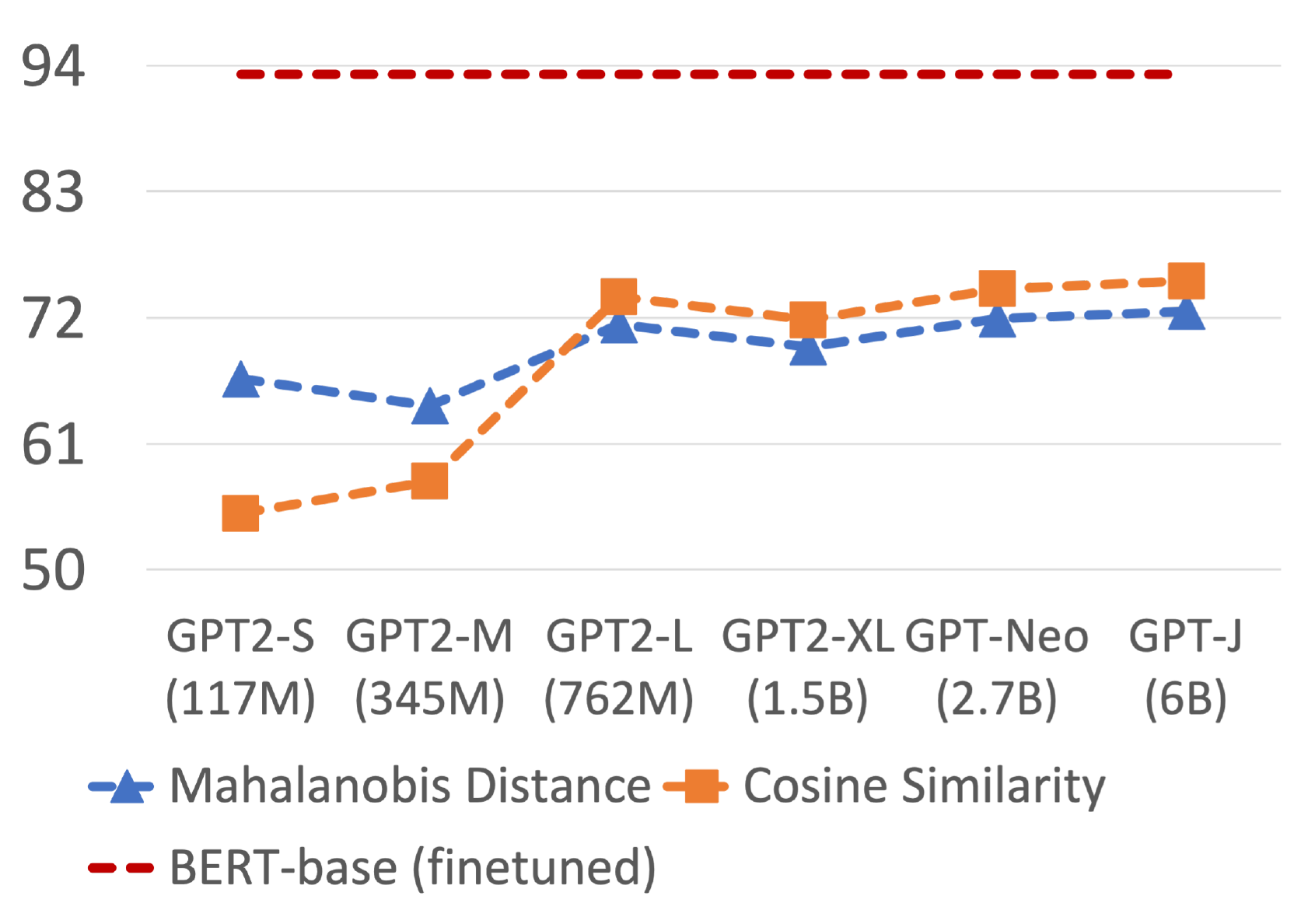}
            \caption{Performance on close-OOD setting.}
            \label{fig:how-zeroshot-close-ood}
        \end{subfigure}
        \caption{OOD detection performance of PLMs without updating the model parameters.}
        \label{fig:how-zeroshot}
    \end{figure*}
    %--- figure end---

        \subsection{OOD Evaluation Methods}
            Evaluation in OOD detection is done via a scoring function, which outputs the appropriateness of the input into a single scalar value ($p$).
            Then we compare $p$ with the pre-set threshold $\delta$ to determine whether the input is an outlier or not:
            \begin{equation}
                \label{eq:MSP-detection}
                I_\delta(\bm{x}) =
                    \left\{
                    	\begin{array}{ll}
                    		\text{IND} &  p(\bm{x}) \ge \delta \\
                    		\text{OOD} &  p(\bm{x}) < \delta,
                    	\end{array}
                    \right
            .
            \end{equation}
            In this paper, we evaluate the performance of our method in 4 different evaluation methods, which can be categorized into 2 higher branches: representation-based and logit-based. \\
        
            \noindent\textbf{Logit-based} approaches exploit the PLM's prediction result extracted from the classification layer as their primary information to discern outliers.
            Logit-based approaches are simple and have their own dominance in computational cost since it pursues OOD detection and general classification nigh simultaneously.
            
            \noindent$\bullet$ \textbf{MSP} is a baseline method in this branch that employs the maximum softmax probability to score the appropriateness of the given input, based on the idea that the model will output more certain output (higher probability) to a normal sample \cite{hendrycks2016baseline}:
            \begin{equation}
                p(\bm{x}) = \frac{e^{f_{i}(\bm{x})}}{\Sigma_{j=1}^{N}e^{f_{j}(\bm{x})}},
            \end{equation}
            \noindent where $f_{i}(\bm{x})$ refer to as max value from the classification layer (max logit value). 
            
            \noindent$\bullet$ \textbf{Energy} is a variant of MSP, which calibrates logit value based on energy function \cite{liu2020energy}:
            \begin{equation}
                p(\bm{x}) = - E(\bm{x};f) = T\cdot\log\Sigma_{i}^{N}e^{f(\bm{x})/T}.
            \end{equation}

            \noindent\textbf{Representation-based} approaches, on the other hand, employ the hidden representation from PLM as their primary source.
            Since the size of the hidden representation is larger and inheres more copious information, they generally yield a more precise decision than logit-based approaches.
            However, they require more inference time to derive a final score.
            We employed Mahalanobis distance-based and cosine similarity-based methods in this branch.
            
            \noindent$\bullet$ \textbf{Mahalanobis distance} refers to the distance between the specific distribution and the input.
            In OOD detection, we estimate the gaussian distribution of the training dataset and utilize the minimum Mahalanobis distance to score the input suitability \cite{lee2018simple}:
                \begin{equation}
                    \label{eq:Mahalanobis-score}
                    p(\bm{x}) =
                    (\bm{h} - \bm{\mu}_k)^\top
                    \Sigma^{-1}
                    (\bm{h} - \bm{\mu}_{k}),
                \end{equation}
                % \vspace{-0.7cm}
            where training distribution is ($\mathcal{N}(\bm{\mu}_i, \Sigma)$ for $i \in i = \{1,2,\cdots, |C|\}$), and $k$ refers to a index of minimum mahalanobis distance. 
                
            \noindent$\bullet$ \textbf{Cosine Similarity} method utilizes the cosine distance between the representation of the given input ($z(x)$) and the nearest neighbor $z(x_{nn})$ \cite{tack2020csi}:
            \begin{equation}
                p(\bm{x})=\text{sim}(\bm{z}(\bm{x}), \bm{z}(\bm{x}_{nn}))
            \end{equation}

    \section{Analysis}
        In this section, we share several intriguing findings and insights from various settings.
        \subsection{OOD Robustness of PLMs without Supervision.}
            In this experiment, we investigate the OOD detection capability of PLMs without parameter tuning. 
            Precisely, we extract the final layer representation from each frozen PLM and evaluate their performance via representation-based evaluation methods.
            (Logit-based evaluation methods are not used as they require additional training of the classification layer.)
            Figure \ref{fig:how-zeroshot} summarizes the results in two scenarios (i.e., far-OOD and close-OOD).
            We verified the correlation between the size of PLMs and their OOD detection ability, but utilizing them without parameter supervision is roughly impossible since they still lag far behind the small supervised methods (i.e., BERT-base with Mahalanobis evaluation) in a barebone setting.
            Moreover, performance improvement from the scaling saturates in a more harsh setting  (i.e., close-OOD), displaying an unbridgeable gap with the fine-tuned model.

\begin{table}[t]
    \centering
    \resizebox{0.48 \textwidth}{!}{%
    \begin{tabular}{c|l|llll}
    \toprule
    \multirow{2}{*}{Setting} & \multicolumn{1}{c|}{\multirow{2}{*}{Backbone}} & \multicolumn{4}{c}{Evaluation Method} \\ \cline{3-6} 
     & \multicolumn{1}{c|}{} & \multicolumn{1}{c|}{MSP} & \multicolumn{1}{c|}{Energy} & \multicolumn{1}{c|}{Mahal.} & \multicolumn{1}{c}{Cosine} \\
     \midrule
    \multirow{6}{*}{\begin{tabular}[c]{@{}c@{}}CLINC\\ Setting\end{tabular}} & GPT2-S & 93.22 & \textbf{95.79} & 77.63 & 76.34 \\ \cline{2-2}
     & GPT2-M & 95.41 & \textbf{97.63} & 82.42 & 79.82 \\ \cline{2-2}
     & GPT2-L & 96.21 & \textbf{97.77} & 96.93 & 97.57 \\ \cline{2-2}
     & GPT2-XL & 96.48 & \textbf{97.99} & 97.28 & 97.66 \\ \cline{2-2}
     & GPT-Neo & 96.04 & \textbf{97.72} & 96.59 & 97.64 \\ \cline{2-2}
     & GPT-J & 97.34 & \textbf{98.50} & 97.91 & 98.20 \\
     \midrule
    \multirow{6}{*}{\begin{tabular}[c]{@{}c@{}}Banking\\ Split 25\%\end{tabular}} & GPT2-S & 90.12 & \textbf{91.32} & 75.32 & 73.11 \\ \cline{2-2}
     & GPT2-M & 91.74 & \textbf{92.78} & 78.03 & 76.56 \\ \cline{2-2}
     & GPT2-L & 93.02 & \textbf{93.45} & 92.44 & 93.41 \\ \cline{2-2}
     & GPT2-XL & 94.29 &\textbf{94.95} & 93.24 & 94.10 \\ \cline{2-2}
     & GPT-Neo & 93.83 & \textbf{94.85} & 92.79 & 93.88 \\ \cline{2-2}
     & GPT-J & 94.11 & \textbf{95.10} & 93.66 & 94.80 \\ 
     \bottomrule
    \end{tabular}}
    \caption{AUROC of each PLMs trained with LoRA. Energey function consistently outperforms other methods .}
    \label{tab:how-evaluation-methods}
\end{table}

    \begin{table*}[t!]
    \centering
    \resizebox{0.9 \textwidth}{!}{%
    \begin{tabular}{c|l|c|llllll}
    \toprule
    \multirow{2}{*}{Setting} & \multicolumn{1}{c|}{\multirow{2}{*}{Method}} & \multicolumn{1}{c|}{\multirow{2}{*}{\# Params.}} & \multicolumn{6}{c}{Backbone} \\ \cline{4-9} 
     & \multicolumn{1}{c|}{} & \multicolumn{1}{c|}{} & \multicolumn{1}{c|}{\begin{tabular}[c]{@{}c@{}}GPT2\\ (S)\end{tabular}} & \multicolumn{1}{c|}{\begin{tabular}[c]{@{}c@{}}GPT2\\ (M)\end{tabular}} & \multicolumn{1}{c|}{\begin{tabular}[c]{@{}c@{}}GPT2\\ (L)\end{tabular}} & \multicolumn{1}{c|}{\begin{tabular}[c]{@{}c@{}}GPT2\\ (XL)\end{tabular}} & \multicolumn{1}{c|}{\begin{tabular}[c]{@{}c@{}}GPT\\ Neo\end{tabular}} & \multicolumn{1}{c}{GPT-J} \\
     \midrule
    \multirow{11}{*}{\begin{tabular}[c]{@{}c@{}}CLINC\\ (far-ood)\end{tabular}} & Linear (SLP) & 0\% & 83.03 & 87.39 & 88.47 & 89.55 & 89.44 & 91.94 \\ \cline{2-9}
     & Fine-tuning & 100\% & 96.84 & 97.71 & 98.24 & 98.33 & 98.01 & 98.41 \\ \cline{2-9}
     & \multirow{3}{*}{LoRA} & 0.1\% & 95.00 & 96.54 & 97.66 & 97.72 & 98.14 & 97.79 \\ \cline{3-3}
     &  & 0.5\% & 96.41 & 96.04 & 97.52 & 97.45 & 98.12 & 97.89 \\ \cline{3-3}
     &  & 1\% & 96.13 & 95.89 & 97.61 & 97.40 & 98.11 & 98.50 \\ \cline{2-9}
     & \multirow{3}{*}{Adapter} & 0.1\% & 96.62 & 97.52& 97.74 & 97.71 & 97.81 & 96.80\\ \cline{3-3}
     &  & 0.5\% & 95.64 & 97.07 & 97.86 & 96.94 & 97.98 & 98.37 \\ \cline{3-3}
     &  & 1\% & 95.79 & 97.63 & 97.77 & 97.99 & 98.12 & 98.50 \\ \cline{2-9}
     & \multirow{3}{*}{Prefix} & 0.1\% & 95.53& 96.93 & 96.38 & 97.88& 90.25 & 98.55 \\ \cline{3-3}
     &  & 0.5\% & 96.91& 96.96 & 97.78 & 97.88 & 89.81 & 97.92 \\ \cline{3-3}
     &  & 1\% & 96.97 & 97.50 & 97.69 & 97.81 & 88.98 & 98.62 \\
     \midrule
     \multirow{11}{*}{\begin{tabular}[c]{@{}c@{}}Banking\\split 25\% \\ (close-ood)\end{tabular}} & Linear (SLP) & 0\% & 72.97 & 75.17 & 80.46 & 77.59 & 86.55 & 89.12 \\ \cline{2-9}
     & Fine-tuning & 100\% & 90.06 & 92.06 & 93.14 & 93.23 & 92.54 & 93.73 \\ \cline{2-9}
     & \multirow{3}{*}{LoRA} & 0.1\% & 91.18 & 91.74 & 94.65 & 94.58 & 94.29 & 95.82 \\ \cline{3-3}
     &  & 0.5\% & 91.16 & 92.98 & 94.54 & 94.04 & 94.55 & 94.65 \\ \cline{3-3}
     &  & 1\% & 91.39 & 92.39 & 93.45 & 93.59 & 94.81 & 95.29 \\ \cline{2-9}
     & \multirow{3}{*}{Adapter} & 0.1\% & 91.97 & 93.24 & 94.90 & 94.69 & 93.26 & 95.59 \\ \cline{3-3}
     &  & 0.5\% & 92.90 & 92.63 & 95.18 & 95.24 & 93.61 & 95.83 \\ \cline{3-3}
     &  & 1\% & 91.32 & 92.78 & 95.41 & 94.95 & 94.41 & 95.37 \\ \cline{2-9}
     & \multirow{3}{*}{Prefix} & 0.1\% & 91.22 & 91.92 & 93.96 & 93.48 & 81.9 & 94.93 \\ \cline{3-3}
     &  & 0.5\% & 91.85 & 92.55 & 93.84 & 93.34 & 80.82 & 93.99 \\ \cline{3-3}
     &  & 1\% & 92.09 & 92.65 & 94.38 & 93.74 & 89.66 & 94.39 \\ 
    \bottomrule
    \end{tabular}
    }
    \caption{AUROC of various PETL methods with various number of parameters evaluated by the energy function.}
    \label{tab:how-numofparams}
    \end{table*}
    
    \begin{figure*}[t]
    \centering
        \begin{subfigure}{.49\linewidth}
            \centering
            \includegraphics[width=.99\linewidth]{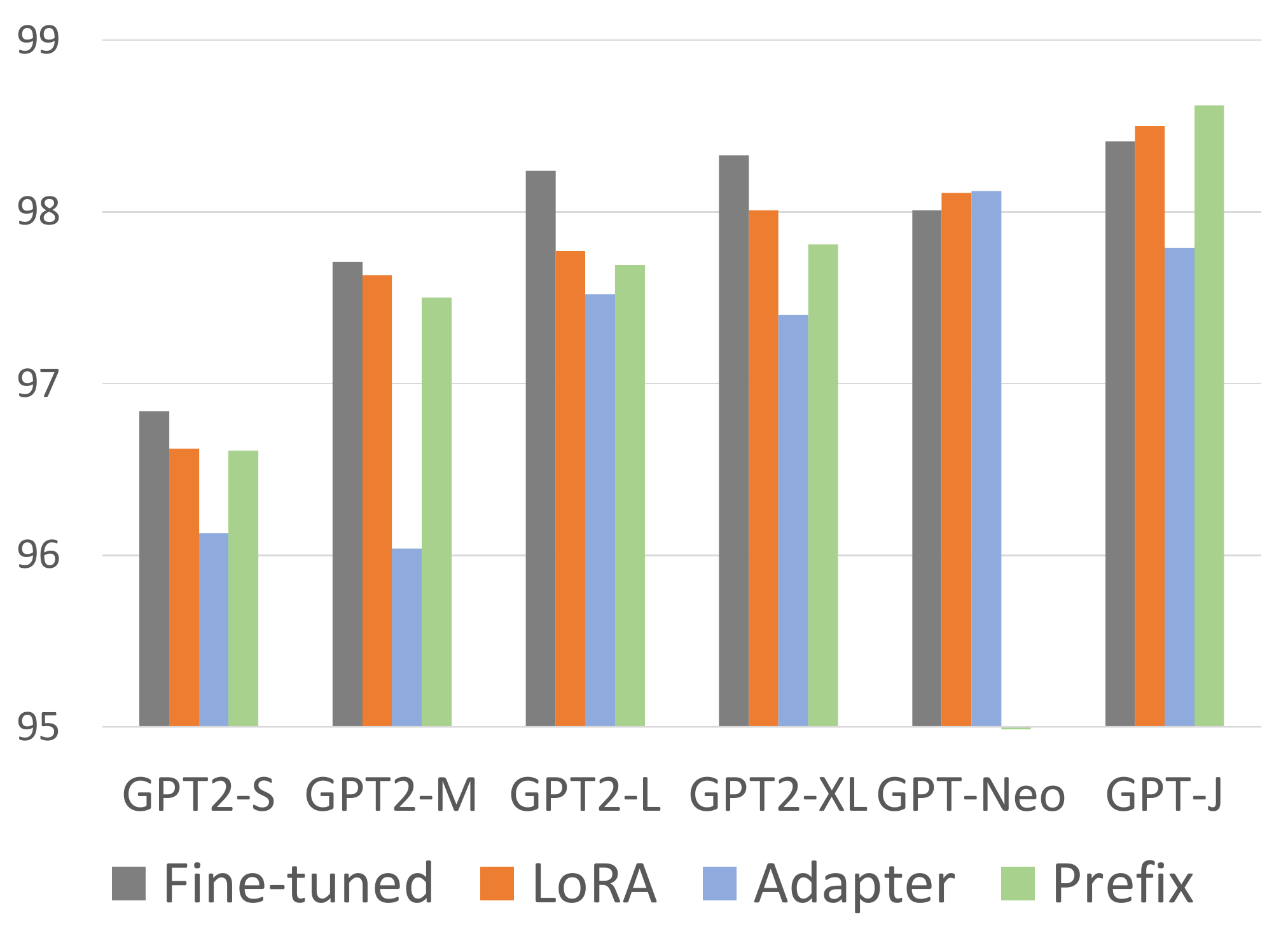}
            \caption{Performance on far-OOD setting.}
            \label{fig:how-PETL-far-ood}
        \end{subfigure}
        \begin{subfigure}{.49\linewidth}
            \centering
              \includegraphics[width=.99\linewidth]{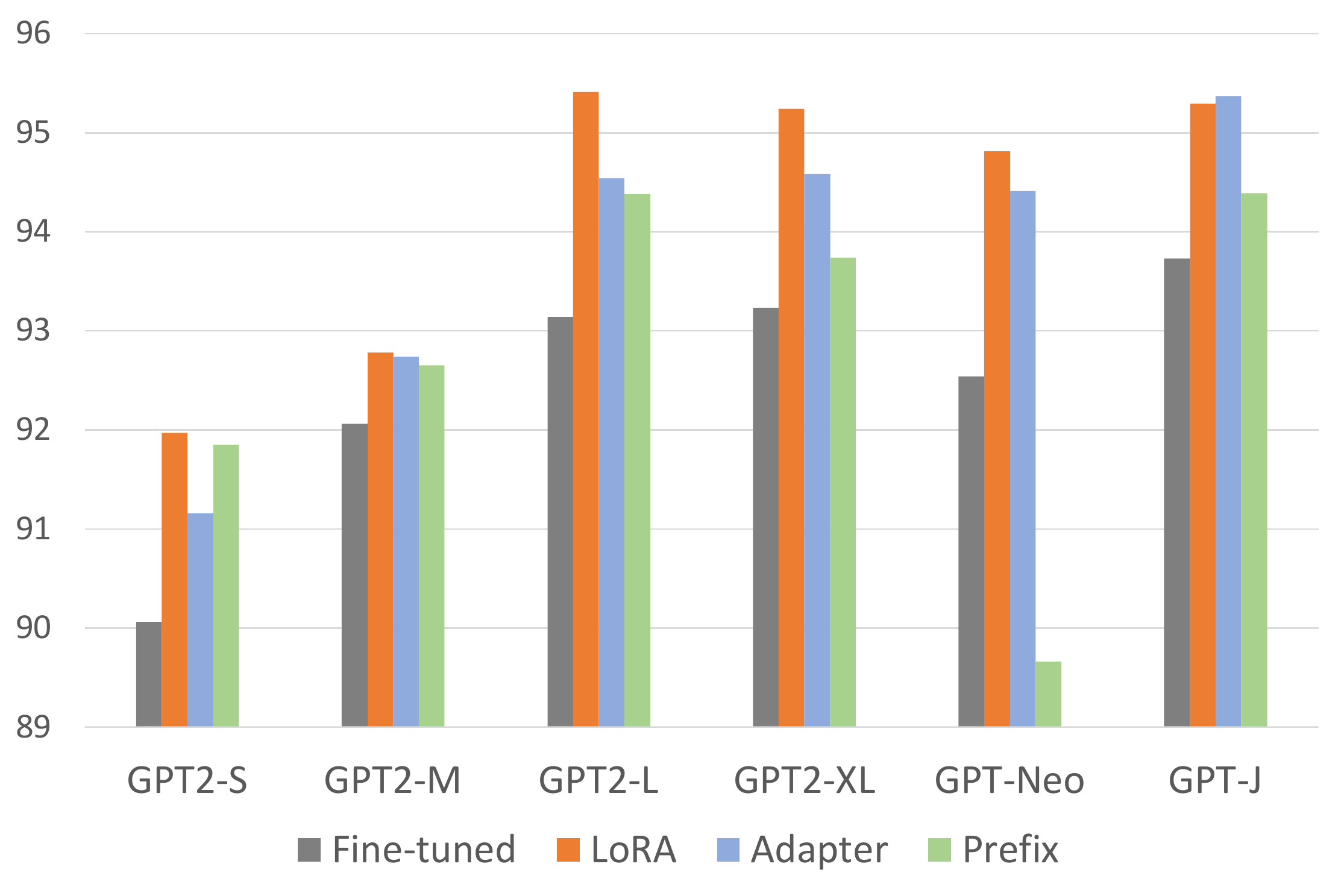}
            \caption{Performance on close-OOD setting.}
            \label{fig:how-PETL-close-ood}
        \end{subfigure}
        \caption{OOD detection performance of PLMs without updating the model parameters.}
        \label{fig:how-PETL}
    \end{figure*}
    %--- figure end---

        \subsection{Evaluation methods for auto-regressive PLMs.}
            Many recent OOD works \cite{zhou2021contrastive,shen2021enhancing}  leverage hidden representation-based evaluation, as they generally surpass logit-based evaluations \cite{podolskiy2021revisiting}.
            The reasonable conjecture behind their success is that hidden representations have more copious information than the logit value.
            However, in auto-regressive PLMs, logit-based evaluations (i.e., MSP and Energy) outperform representation-based methods (i.e., Mahalanobis distance and cosine similarity), as shown in Table \ref{tab:how-evaluation-methods}.
            The reasonable conjecture for this phenomenon is due to the characteristic of the language model.
            Unlike bi-directional models (e.g., BERT, RoBERTa, DeBERTa), decoder models (e.g., GPT and its variants) do not have [CLS] embedding, which assembles the token embeddings to capture holistic information \cite{devlin2018bert,kim2021self}.
            Therefore, auto-regressive PLMs generally utilize the last token embedding as a final feature embedding replacing [CLS] embedding of encoder-based models.
            While the last token of GPT is befitted for predicting the next token, however, it cannot extract the holistic semantics of the sentence suitably, unlike [CLS] embedding.
            We believe extracting a better representation through various pooling \cite{wang2020sbert} methods might be a possible avenue for auto-regressive models to improve the OOD robustness further.
        
        \subsection{PETLs VS. Fine-tuning}
            In this experiment, we investigate the performance gap between various PETL methods (i.e., Adapter, LoRA, prefix-tuning) and model fine-tuning.
            To compare the performance of each method under similar circumstances, we set every PETL method to utilize a similar number of parameters sufficient enough to reach maximum accuracy. 
            % (Additional experiments are detailed in next experiment.)
            Moreover, we utilized the energy function to evaluate each method as they displayed the best performance among other evaluation methods, i.e., cosine, Mahalanobis, and MSP, in the previous experiments.
            Table \ref{tab:how-numofparams} summarizes the results.
            
            From this experiment, we observed that PETL methods are more robust than fine-tuning with reasonably large PLMs (i.e., GPT-J).
            Specifically, most PELT methods on GPT-J outperform fine-tuning with proper tunable parameters.
            Nevertheless, size is not the ultimate answer.
            While it is clear that the scale of a model is an essential factor in OOD robustness, larger models are still vulnerable to close-OOD inputs.
            The capability to detect far-OOD inputs (far from the training distribution) improves proportionally as the size grows, while the ability to identify close-OOD input improves rather trivially.
            PLM's vulnerability to close-OOD has already been reported in other studies \cite{zhang2021pretrained}, and this may be related to shortcut learning \cite{geirhos2020shortcut} that predicts with high probability by looking at specific words.
            Generating OOD data with particular keywords or utilizing another pretext task, such as \cite{moon2020masker}, can be worthy approaches to alleviate such phenomena.
            A suitable OOD approach is necessary to alleviate the aforementioned issue, as it can further boost the robustness.
            We conduct additional experiments with PETLs on three different numbers of tunable parameters: 0.1\%, 0.5\%, and 1\% of the PLM parameters.
            Figure \ref{tab:how-numofparams} summarizes the results.
            With sufficient parameters to reach maximum performance, there is no meaningful difference or improvement within each methodology.
            Also, empirically, we confirmed that LoRA is the most stable during learning and that prefix-tuning fluctuates severely according to learning.
    
    \section{Conclusion and Future Work}
        In this study, we showed that the scale of the language model is an important factor in OOD robustness. Moreover, we also showed that various methodologies outperform fine-tuning when applied to sufficiently large PLM. 
        Our follow-up work seeks to create a methodology that allows large PLMs to be more robust to OOD input.
        The performance improvement that can be achieved by the size of PLM and OOD technique is orthogonal.
        In line with the growing size of PLM, the OOD technique needs to be developed in a more parameter-efficient way.
        As such, developing a proper OOD technique compatible with the parameter-efficient transfer methods is our proper goal.

    \bibliography{anthology,custom}
    \bibliographystyle{acl_natbib}
    \clearpage
    
    \appendix
    \label{sec:appendix}
    \noindent{\Large\bfseries Appendix}

    \section{Related Work}
        \noindent\textbf{Parameter-Efficient Transfer Learning} is drawing considerable attention lately, emerging as an alternative strategy to fine-tuning.
        Compared to fine-tuning, parameter-efficient transfer methods show superiority in the number of trainable parameter usage while achieving performance analogous to fine-tuning.
        Depending on the characteristics of the methods, parameter-efficient transfer methods can be categorized into \textit{Adapter-based} and \textit{Prompt-Engineering} approaches.

        \textit{Adapter} \cite{houlsby2019parameter, pfeiffer2020adapterfusion} refers to a lightweight neural module injected within each layer of PLM.
        The structure of the adapter generally consists of a bottleneck layer (down-projection and up-projection), a nonlinear function, a normalization layer, and a residual connection.
        The adapter has many different variants due to numerous design choices, such as the order or specifics of each component (e.g., which normalization technique will be used) and where the adapter will be attached.
        For example, LoRA \cite{hu2021lora} inserts low-rank decomposition matrices in each weight in self-attention \cite{vaswani2017attention} (i.e., query, key, and value).
        
        Another line of work, \textit{prompt engineering}, casts the existing task as a text generation problem to fully leverage the capability of PLMs to predict the appropriate word in the given sentence.
        This approach requires an empirical endeavor of optimizing the prompt to maximize a PLM’s performance.
        Earlier works exploit handcrafted manual prompts \cite{schick2020exploiting, jiang2020can} or by providing demonstrations to PLM \footnote{also termed as in-context learning.} \cite{brown2020language, raffel2019exploring, gao2020making, zhao2021calibrate}.
        More recent work replaces the manual prompt with a soft prompt \cite{li2021prefix, lester2021power, shin2020autoprompt, liu2021gpt}, a machine trainable continuous vector.
        The soft prompt is a more modular and versatile method that evades additional latency in the inference phase because it detaches the additionally trained parameters and solely employs the final output of the trained parameters as the prompt.

        While former parameter-efficient transfer methods showed noticeable achievements, their evaluations generally assume the train and test distributions are identical (i.e.,  i.i.d. assumption);
        however, this condition is rarely satisfied in real-world scenarios due to the diversity and volatility of user input.
        Consequently, if the model can not correctly handle distribution-shifted malicious input and misconceives it as an in-distribution (IND) example, it may lead to fatal accidents.
        
        Despite its practical importance, how large PLMs or parameter-efficient transfer learning cope with unknown input is poorly understood.
        This work aims to understand language models' capabilities to detect outliers through parameter-efficient transfer learning methods.

    \section{Parameter-Efficient Transfer Learning}
        \noindent\textbf{Adapter}
        The adapter approach inserts small trainable adapter modules between transformer layers while the parameters of the original network remain fixed. The adapter module uses a bottleneck architecture which projects the input dimension $h$ to a lower-dimensional space specified by bottleneck dimension $r$, followed by a nonlinear activation function, and a up-projection to initial dimension $h$. In this work, we attach adapter modules in two places, i.e., after the projection following multi-head attention and after the two feed-forward layers, following original implementation in \cite{houlsby2019parameter}.
        Also, we use relu as a nonlinear function and layer normalization \cite{ba2016layer}.
        
        \noindent\textbf{LoRA}
        LoRA injects trainable low-rank matrices into transformer layers to approximate the weight updates. For a pre-trained weight matrix $W \in {\mathbb{R}^{h\times{k}}}$, LoRA decompose $\Delta W=W_{down}W_{up}$ where $W_{down} \in {\mathbb{R}^{h\times{r}}}, W_{up} \in {\mathbb{R}^{r\times{k}}}$ are trainable parameters. Specifically we attach LoRA in weight matrices in the self attention module. 
        Specifically we attached LoRA to query and key vector following the original implementation.
        
        \noindent\textbf{Prefix-Tuning}
        Prefix tuning prepends $l$ tunable prefix vectors to the keys and values of the multi-head attention at every layer. Following the original implementation, we reparametrize the prefix matrix of dimension $h$ by a smaller matrix of dimension $r$ composed with a large feedforward neural network with tanh as a nonlinear function.

    \section{Expanded Configuration Details}
        \subsection{Common Environment}
            For the experiments, 4 Tesla V100 SXM2 32GB GPUs are used. The batch size is 8 per GPU. When the GPU is too small for the batch size, we set batch size to 4 and the number of gradient accumulation steps to 2.
            We implemented our model based on Transformers \cite{wolf-etal-2020-transformers} library by Huggingface. Additionally, we used deepspeed \cite{rajbhandari2020zero} to train models. Specifically, we used ZeRO2 with cpu offload on a 240GB RAM CPU. In this setting, fine-tuning GPT-J on CLINC150 full dataset takes about 7.1 GPU hours per epoch. We used AdamW \cite{loshchilov2017decoupled} optimizer with epsilon 1e-6 and weight decay 0.1. Furthermore, we apply the cosine annealing scheduler. For GPT-neo, the minimum learning rate is 0. For GPT-J, the  minimum learning rate is the one fifth of maximum learning rate.

    \begin{table}[t]
    \centering
    \resizebox{\columnwidth}{!}{%
    \begin{tabular}{l|c|c|c|c|c}
    \toprule
    dataset & \#domain & \#intent & \#data (train/val/test/ood)\\
    \midrule
    CLINC & 10 & 15  & 15000/3000/4500/1000\\
    Banking& 1 & 77  & 7812 / 1520 / 3040\\
    \bottomrule
    \end{tabular}
    }
    \caption{Dataset statistics.}
    \label{tab:datasets}
    \vspace{-0.3cm}
    \end{table}
    
    \begin{table}[t!]
        \centering
        \resizebox{\columnwidth}{!}{%
            \begin{tabular}{l|ccc}
                \toprule
                \multirow{2}{*}{BERT-base} & \multicolumn{3}{c}{CLINC150 Full} \\ \cline{2-4}
                     & \multicolumn{1}{c|}{ACC $\uparrow$} & \multicolumn{1}{c|}{FPR-95 $\downarrow$} & \multicolumn{1}{c}{AUROC $\uparrow$}\\ \hline
                    \citet{shu2017doc} & 94.51{\footnotesize $\pm0.45$} & 23.33{\footnotesize $\pm1.27$} & 95.92{\footnotesize $\pm0.05$} 
    \\
                    
                    \citet{li2021cross} & 96.1{\footnotesize $\pm0.37$} & {10.6{\footnotesize $\pm0.26$}} & {97.72{\footnotesize $\pm0.03$}}
     \\
                    
                    \citet{zeng2021modeling}  & 94.19{\footnotesize $\pm0.28$} & 23.4{\footnotesize $\pm1.97$} & 95.75{\footnotesize $\pm0.2$}
     \\
                    \citet{ zhou2021contrastive} & 95.79{\footnotesize $\pm0.13$} & 10.7{\footnotesize $\pm0.95$} & 97.6{\footnotesize $\pm0.11$}
    \\
                    \citet{shen2021enhancing} & 96.66 & 10.88 & 97.43  
    \\
                    \citet{cho2022enhancing} & \textbf{96.96}{\footnotesize $\pm0.39$} & \textbf{6.67} {\footnotesize $\pm0.51$} & \textbf{98.27} {\footnotesize $\pm0.16$} 
    \\
                \bottomrule
            \end{tabular}
        }
        \caption{Results of each model trained on the CLINC150 dataset. The best performance in each metric is indicated in \textbf{bold}. }
        \label{tab:bert-baseline}
    \end{table}

    \begingroup
    \setlength{\tabcolsep}{10pt} % Default value: 6pt
    \begin{table*}[t]
    \resizebox{\textwidth}{!}{%
    \begin{tabular}{l|l|c}
    \toprule
        Method & Parameters & Values \\
    \midrule
        \multirow{3}{*}{LoRA} & Learning rate & 2e-4 (GPT-Neo), 5e-5 (GPT-J) \\ 
         & Bottleneck dim & 8 (GPT-Neo / 0.1\%), 80 (GPT-Neo / 1\%), 12 (GPT-J /0.1\%), 128 (GPT-J / 1\%) \\
         & Location & query, value \\
    \midrule
        \multirow{3}{*}{Adapter} & Learning rate & 8e-5 (GPT-Neo / 0.1\%), 1e-4 (GPT-Neo / 1\%), 5e-5 (GPT-J), 5e-4 (GPT-J / 0.1\%), 1e-4 (GPT-J / 1\%) \\
         & Bottleneck dim &  6 (GPT-Neo / 0.1\%), 80 (GPT-Neo / 1\%), 11 (GPT-J /0.1\%), 128 (GPT-J / 1\%)\\
         & Location & after Multi-head, after Feed-forward, \\
    \midrule
        \multirow{3}{*}{Prefix-tuning} & Learning rate & 2E-4 (GPT-Neo), 5E-5 (GPT-J) \\ 
         & Bottleneck dim &  12 (GPT-Neo / 0.1\%), 160 (GPT-Neo / 1\%), 20 (GPT-J /0.1\%), 256 (GPT-J / 1\%) \\
         & Prefix length & 5, 10, 20\\
    \bottomrule
    \end{tabular}%
    }
    \caption{Hyper-parameter search for each model.}
    \label{hyperparameter}
    \end{table*}
    \endgroup

        \subsection{Number of Trainable-Parameter}
            For each method, a feed-forward layer is added at the end of the model. In this section, we will calculate the number of additional trainable parameters of each training methods discussed in this paper. Biases are omitted for better readability.
            
            \noindent\textbf{Adapter} Adapter method adds four feed-forward layers per transformer layer in the model. Two of them are down-projection layers, and the others are up-projection layers. When the original embedding size of the model is $h$, the bottleneck dimension is $r$, and the number of transformer layers is L, the number of the trainable parameters of these layers is calculated as $4Lhr$, excluding the bias of the added layers.
            
            \noindent\textbf{LoRA} Similar to adapter, LoRA also adds feed-forward layers per transformer layer. Therefore, the number of the trainable parameters of $4Lhr$. However, the number of parameters are less than adapter if $h$ and $r$ is the same, since LoRA does not use bias of the feed-forward layers.
            
            \noindent\textbf{Prefix-Tuning} There are two trainable elements in prefix tuning. The first one is the prefix embeddings. When the number of prefixes is $l$, and the embedding size is $h$, $lh$ parameters are used by the prefixes. Second, the reparametrization matrix is also trained. The down-projection matrix has $hr$ parameters, when the reduced dimension for reparametrization is $r$. The up-projection matrix has $2Lhr$ parameters. As a result, there are $h(2Lr + l)$ trainable parameters on prefix tuning approach.  
        
    \subsection{Hyper-parameter Search}
        
        Tab \ref{hyperparameter} summarizes hyper parameters for each model.

    \section{Selecting SOTA OOD Method.}
        The Tab.\ref{tab:bert-baseline} summarizes the results with recently proposed OOD approaches on BERT-base with CLINC dataset. 
        The best performing model \cite{cho2022enhancing} is selected as the baseline.

\end{document}